\pdfoutput=1

\documentclass[11pt]{article}

\usepackage[final]{acl}

\usepackage{times}
\usepackage{latexsym}

\usepackage[T1]{fontenc}

\usepackage[utf8]{inputenc}

\usepackage{microtype}

\usepackage{inconsolata}

\usepackage{amsmath}
\usepackage{makecell}
\usepackage{adjustbox}
\usepackage{multicol,multirow}
\usepackage{subfigure}
\usepackage{tabularx}
\usepackage{hyperref}
\usepackage{url}

\usepackage{bm}  
\usepackage{amssymb}  
\usepackage{bbm}  
\usepackage{xcolor}  
\usepackage{wrapfig}
\usepackage{float}
\usepackage{algorithm}
\usepackage{algorithmic}
\usepackage{booktabs}
\usepackage{textcomp}

\title{Knowledge Distillation of Black-Box Large Language Models}

\author{
Hongzhan Chen\textsuperscript{1}, Runjun Chen \textsuperscript{1}, Yuqi Yi \textsuperscript{1}, Xiaojun Quan\textsuperscript{1}\thanks{\; Corresponding author.}, \\
\textbf{Chenliang Li\textsuperscript{2}, Ming Yan\textsuperscript{2} and Ji Zhang\textsuperscript{2}} \\
  \textsuperscript{1}School of Computer Science and Engineering, Sun Yat-sen University, China \\
  \textsuperscript{2}Alibaba Group, China \\
  \textsuperscript{1}\texttt{chenhzh59@mail2.sysu.edu.cn, quanxj3@mail.sysu.edu.cn}\\
  \textsuperscript{2}\texttt{ym119608@alibaba-inc.com}
} 

\begin{document}
\maketitle
\begin{abstract} 
Given the exceptional performance of proprietary large language models (LLMs) like GPT-4, recent research has increasingly focused on boosting the capabilities of smaller models through knowledge distillation (KD) from these powerful yet black-box teachers. While leveraging the high-quality outputs of these teachers is advantageous, the inaccessibility of their internal states often limits effective knowledge transfer. To overcome this limitation, we introduce Proxy-KD, a novel method that uses a proxy model to facilitate the efficient transfer of knowledge from black-box LLMs to smaller models. Our experiments show that Proxy-KD not only enhances the performance of KD from black-box teacher models but also surpasses traditional white-box KD techniques.~This approach presents a compelling new avenue for distilling knowledge from advanced LLMs.
\end{abstract}

\section{Introduction}

Recently, proprietary large language models (LLMs) like GPT-3.5 \citep{openai2022chatgpt} and GPT-4 \citep{openai2023gpt-4} have demonstrated significant superiority over open-source counterparts such as the Llama series \citep{llama,llama2,llama3}. However, their vast number of parameters leads to high inference costs, and they are only accessible via API calls, offering limited customization and transparency. To address these challenges, recent efforts like Alpaca \citep{alpaca}, Vicuna \citep{vicuna2023}, and Orca \citep{mukherjee2023orca} have focused on transferring the capabilities of proprietary LLMs to smaller open-source models through knowledge distillation \citep{chen-etal-2023-mcc,hsieh-etal-2023-distilling,ho2022reasoning-teacher}.


\begin{figure}[t]
    \centering
    \vspace{0mm}
    \begin{adjustbox}{width=0.48\textwidth}
        \includegraphics{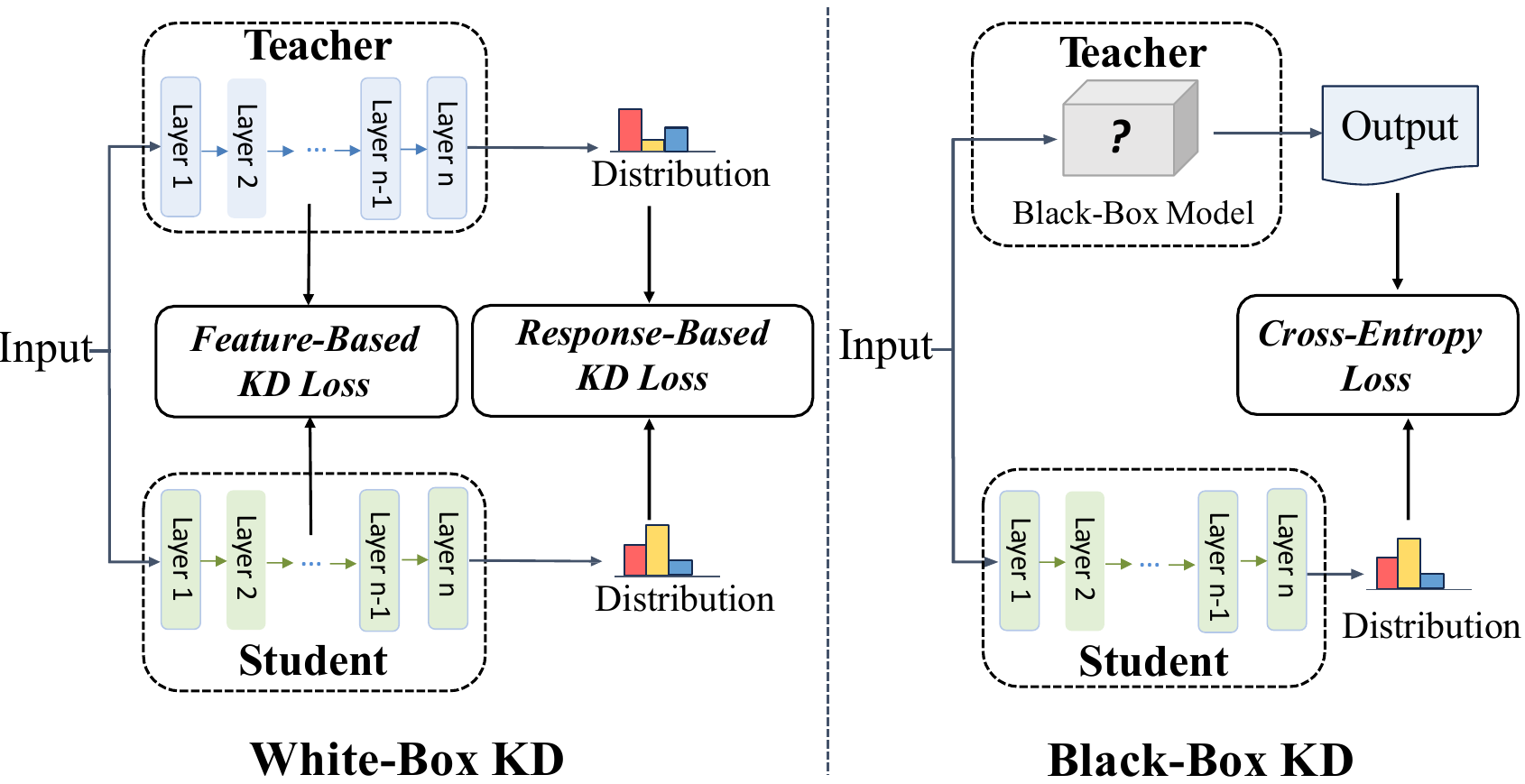}
    \end{adjustbox}
    \vspace{-4mm}
    \caption{Comparison of white-box knowledge distillation (KD) and black-box knowledge distillation (KD).}
    \vspace{-4mm}
    \label{fig:kd}
\end{figure}

Knowledge distillation (KD) \citep{hinton2015distilling} is a technique used to enhance the performance of a smaller student model by learning from a larger, more sophisticated teacher model. Depending on the level of access to the teacher model's internals, KD methods can be categorized into two types: KD with black-box teachers and KD with white-box teachers. As illustrated in Figure \ref{fig:kd}, white-box KD allows the student model to distill more intrinsic knowledge from the teacher by mimicing the teacher model's output distribution \citep{gu2023minillm,wen-etal-2023-f}, hidden states \citep{jiao-etal-2020-tinybert,sun-etal-2019-patient}, and attention scores \citep{wang-etal-2021-minilmv2}. Therefore, this method can only be applied when the teacher model's parameters are accessible. On the other hand, black-box KD leverages the high-quality outputs from powerful proprietary LLMs to fine-tune the student model \citep{hsieh-etal-2023-distilling,fu2023specializing}. Both white-box and black-box KD have their respective drawbacks. While white-box KD is hindered by the limited capacity of the teacher model, which often restricts the distillation performance of the student, black-box KD faces challenges with knowledge transfer due to the inaccessibility of the teacher model's output distribution and internal states.

In this paper, we propose Proxy-based Knowledge Distillation (Proxy-KD) to better transfer knowledge from black-box teacher models. Proxy-KD introduces a proxy model, typically a white-box LLM, between the student and the black-box teacher.
The proxy model first aligns with the capabilities of the black-box teacher by leveraging the teacher's outputs. Moreover, preference optimization is performed to further refine and enhance the alignment between the proxy and teacher models.



During the knowledge distillation process, the proxy model generates a dense distribution that closely approximates the black-box teacher’s output distribution. This enables the student model to train effectively as if it were using the black-box teacher's guidance. To further improve the student's learning effect, we propose incorporating a sample-level weight into the distillation objective. This weight reflects the quality of alignment between the proxy and the teacher model for each sample, allowing the student to concentrate on learning well-aligned distributions from the proxy. 
Moreover, the outputs from the black-box teacher serve as pseudo-labels for the supervised fine-tuning of the student model, akin to traditional white-box knowledge distillation. 
Introducing the proxy model also mitigates the model capacity gap issue \citep{cho2019efficacy}, which typically occurs when there is a notable disparity in capabilities between the teacher and the student.

To validate the effectiveness of our method, we conducted comprehensive experiments across a range of well-established benchmarks. The results show that Proxy-KD consistently outperforms both black-box and white-box KD methods. We observed that the alignment between the proxy model and the black-box teacher is crucial; a poorly aligned proxy model significantly diminishes the performance of knowledge distillation. We also found that larger and more robust proxy models are generally more desirable, as they possess stronger foundational capabilities and can align more effectively with the black-box teacher, enhancing the distillation process. Furthermore, we discovered that directly fine-tuning the proxy model with outputs from the black-box teacher is suboptimal for the alignment, requiring more effective alignment methods. These findings highlight the importance of selecting a well-aligned and capable proxy model to fully leverage the benefits of Proxy-KD. 

We summarize our contribution as below:
\begin{itemize}
\item To tackle the challenge of knowledge distillation for closed-source LLMs, we propose Proxy-KD, which introduces an aligned proxy between the teacher and student models.
    \item 
    We propose a DPO-based alignment strategy for the proxy to align with the teacher and demonstrate that this alignment is essential for Proxy-KD to achieve effective distillation.
   \item We propose to include a sample-level weight in the distillation objective. This weight allows the student to concentrate on learning well-aligned distributions from the proxy.
\end{itemize}

\section{Related Work}

Existing knowledge distillation methods can be categorized into \emph{white-box knowledge distillation} and \emph{black-box knowledge distillation}.

\subsection{White-Box Knowledge Distillation}

Traditional knowledge distillation (KD) research predominantly employs white-box teachers and is typically classified into three main branches: feature-based, response-based, and relation-based methods. Feature-based methods seek to replicate the teacher’s intermediate representations, such as attention scores \citep{jiao-etal-2020-tinybert}, attribution maps \citep{wu-etal-2023-ad}, and hidden representations of tokens \citep{sun-etal-2019-patient}. Response-based methods train the student model by minimizing divergences like Kullback–Leibler (KL) divergence  \citep{hinton2015distilling,sanh2019distilbert}, reverse KL \citep{gu2023minillm,wen-etal-2023-f}, Jensen–Shannon Divergence (JSD) \citep{fang2021mosaicking,yin2020dreaming}, and Total Variation Distance (TVD) \citep{wen-etal-2023-f} based on the teacher’s output distribution. Relation-based methods train the student model by learning pairwise distances and triple-wise angles among token representations from the teacher \citep{park-etal-2021-distilling}, or extracting structural relations from multi-granularity representations \citep{liu-etal-2022-multi-granularity}.

\begin{figure*}
    \centering
    \vspace{-5mm}
    \begin{adjustbox}{width=0.95\textwidth}
        \includegraphics{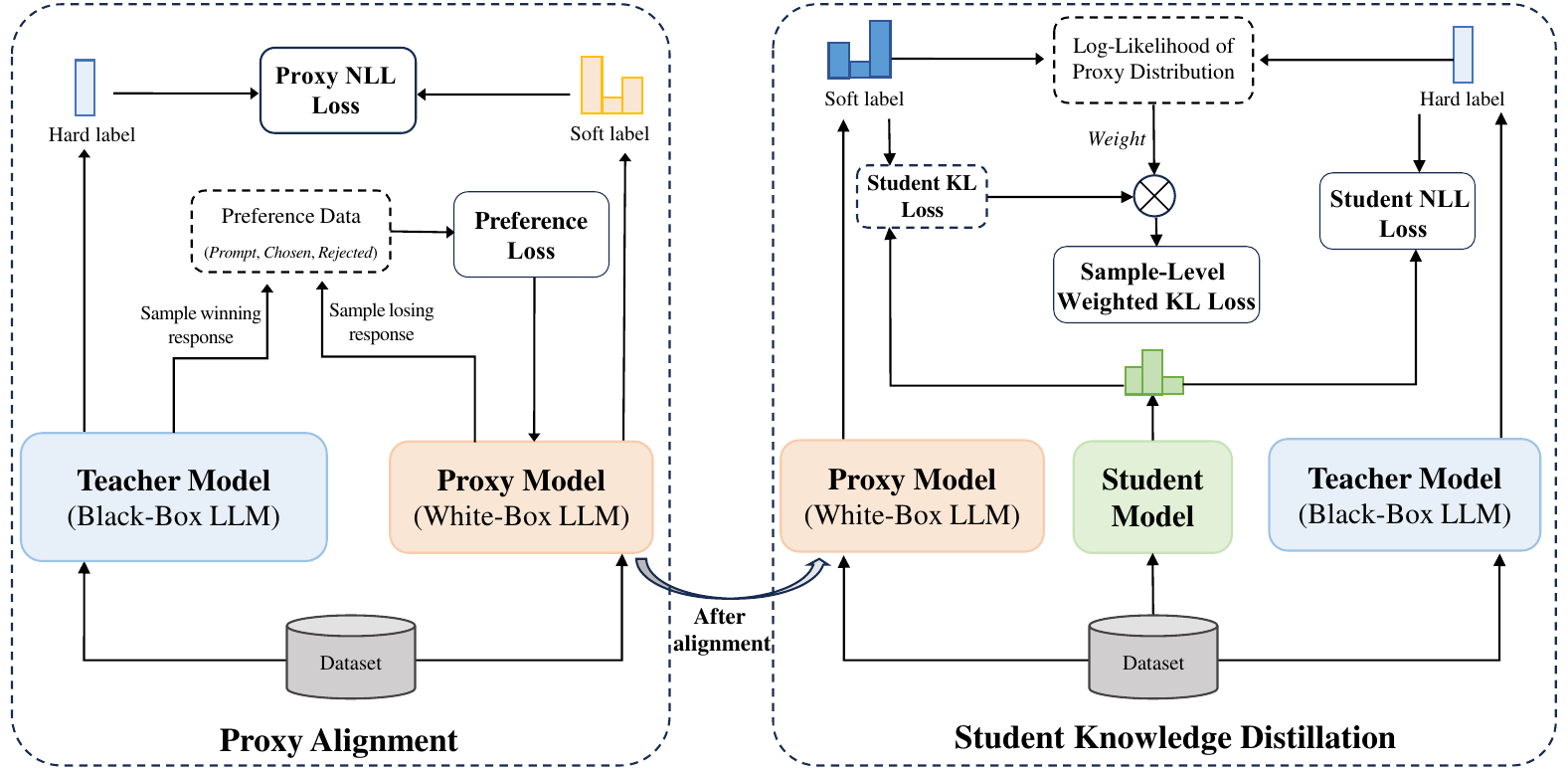}
    \end{adjustbox}
    \vspace{-0mm}
    \caption{Overview of our proposed Proxy-based Knowledge Distillation (Proxy-KD).}
    \vspace{-0mm}
    \label{fig:method}
\end{figure*}

\subsection{Black-Box Knowledge Distillation}

Given the remarkable performance achieved by proprietary LLMs like GPT-4 \citep{openai2023gpt-4}, Claude 3 \cite{anthropic2024claude3}, and Gemini \citep{team2023gemini}, recent studies like Alpaca \citep{alpaca}, Vicuna \citep{vicuna2023}, and Orca \citep{mukherjee2023orca} have focused on  transferring diverse capabilities from these black-box teachers into smaller open-source models. For instance, \citet{li2024common} and \citet{liu2023tinygsm} improved the mathematical capability of small models by training on tailored rationale samples generated by GPT-3.5-Turbo and GPT-4. To transfer the code generation capability, \citet{azerbayev2023explicit} prompted Codex \citep{chen2021evaluating} to create natural language-code pairs and fine-tuned a smaller model on these samples. To transfer the tool usage capability, \citet{gou2023tora} utilized GPT-4 to generate interactive tool-use trajectories as training samples for the target model. Other approaches, such as \citet{hsieh-etal-2023-distilling,ho2022reasoning-teacher,chen-etal-2023-mcc}, utilize rationales generated by black-box teachers as training data to transfer their general reasoning capabilities.

White-box knowledge distillation (KD) efficiently distills knowledge by leveraging the internal states of the teacher model. However, white-box teachers typically possess a more limited capacity compared to their black-box counterparts. In contrast, black-box KD capitalizes on the superior performance of the teacher models but is restricted to fine-tuning on teacher-generated samples. This approach captures input-output patterns without accessing the deeper, intrinsic knowledge of the teacher model. To bridge these gaps, we propose Proxy-KD, a straightforward method that combines the strengths of both white-box and black-box KD while mitigating their respective limitations.

\subsection{Connection with Teacher Assistant}

The proposed Proxy-KD method draws inspiration from TAKD \citep{mirzadeh2019improved}, as both methods use an intermediate network to aid knowledge distillation, but they differ in three significant ways. 
Firstly, the motivation behind each approach is distinct: TAKD focuses on mitigating the capacity gap between the teacher and student in white-box settings, whereas Proxy-KD addresses the challenges posed by black-box teacher models and seeks to incorporate the benefits found in white-box scenarios. 
Secondly, the methodologies diverge, with Proxy-KD introducing a crucial proxy alignment phase that includes preference optimization to better align the proxy model with the black-box LLM. This step is essential for reducing discrepancies between the proxy and teacher models, thereby improving the effectiveness of the distillation process.
Lastly, they operate in different domains: TAKD is applied in the field of computer vision, while Proxy-KD is specifically designed for natural language processing, targeting the distillation of proprietary large language models (LLMs).

\section{Method}

In this section, we introduce Proxy-based Knowledge Distillation (Proxy-KD), a simple yet efficient approach for knowledge distillation from black-box LLMs. As illustrated in Figure \ref{fig:method}, Proxy-KD introduces a larger white-box LLM as the proxy aiming to capture the black-box teacher's knowledge. The process unfolds in two main stages: (1) proxy model alignment and (2) student knowledge distillation. First, the proxy model is aligned with the teacher through supervised fine-tuning and preference optimization. Once aligned, the student model learns from both the explicit outputs (hard labels) of the black-box teacher and output distributions (soft labels) provided by the aligned proxy.



\subsection{Problem Statement}

To facilitate the transfer of knowledge from a black-box teacher LLM $\pi_t$ to a smaller, open-source student LLM $\pi_s$, we introduce a proxy model $\pi_p$. The training dataset $\mathcal{D}$ consists of input-output pairs $(x, y)$, where $x$ represents the input prompt and $y$ is the output sequence generated by the teacher model $\pi_t$. This dataset is strategically divided into three parts: 10\% ($\mathcal{D}_w$) for the warm-up phase, 45\% ($\mathcal{D}_p$) for aligning the proxy model with the teacher, and the remaining 45\% ($\mathcal{D}_s$) for the knowledge distillation training of the student model.

The process begins with a warm-up phase where the proxy model $\pi_p$ is trained on $\mathcal{D}_w$. This phase helps $\pi_p$ develop a basic capability to generate responses to input prompts. Following this, the proxy model undergoes alignment with the teacher model $\pi_t$ using the next dataset, $\mathcal{D}_p$. This alignment is achieved through two methods: hard-label knowledge distillation (KD) and preference learning. These methods enable $\pi_p$ to approximate the behavior and outputs of the teacher model. Once aligned, $\pi_p$ acts as an intermediary, facilitating the transfer of knowledge to the student $\pi_s$ on $\mathcal{D}_s$.





\subsection{Preliminary}


\paragraph{Hard-Label Knowledge Distillation.} In this approach, the student model is trained using the outputs generated by the teacher model by minimizing the negative log-likelihood (NLL) function:
\begin{equation}
\mathcal{L}_{\text{NLL}}=\mathbbm{E}_{(x,y)\sim\mathcal{D}}\left[-\log\pi_s(y|x)\right],
\label{eq:hardlabelKL}
\end{equation}
where $\pi_s(y|x)$ is the probability of $\pi_s$ generating $y$ given $x$. 
This approach is essentially a form of supervised fine-tuning and typically employed when the teacher is a black-box model.

\paragraph{Soft-Label Knowledge Distillation.} In this approach, the student is trained to imitate the token-level probabilities of the teacher, by minimizing the Kullback-Leibler  (KL) divergence:
\begin{equation}
\mathcal{L}_{\text{KL}}=\mathbbm{E}_{(x,y)\sim\mathcal{D}}\left[\mathbb{D}_{\text{KL}}(\pi_t(y|x)||\pi_s(y|x))\right].
\label{eq:softlabelKL}
\end{equation}
This knowledge distillation approach is typically employed when the teacher is a white-box model.

While the KL divergence objective provides richer supervision signals by using the token-level output distributions of the teacher model, it cannot be applied to black-box teachers due to the inaccessibility of these distributions. Consequently, current methods \citep{vicuna2023,mukherjee2023orca} rely on supervised fine-tuning using the outputs generated by black-box models to transfer their knowledge.
Proxy-KD addresses this limitation by using a proxy model to incorporate the KL objective. The proxy mimics the black-box teacher, allowing access to its output distributions and enabling a more effective knowledge transfer.



\subsection{Proxy Model Alignment}
The proxy model $\pi_p$ is typically a larger white-box LLM than the student model $\pi_s$. For effective knowledge transfer, it's crucial to first align the output distribution of the proxy model with that of the black-box teacher model $\pi_t$. This alignment ensures that the proxy accurately captures the teacher’s behavior.

The proxy model $\pi_p$ first undergoes supervised fine-tuning on a warm-up dataset $\mathcal{D}_w$. Following this, the proxy is further trained on the $\mathcal{D}_p$ dataset by minimizing the NLL loss:
\begin{equation}
    \mathcal{L}_{\text{Proxy-NLL}}=\mathbbm{E}_{(x,y)\sim\mathcal{D}_p}\left[-\log\pi_p(y|x)\right].
\end{equation}

To enhance the alignment of the proxy model with the teacher, we further introduce a preference learning-based alignment objective, with the hypothesis that the teacher model's responses are of higher quality compared to those from the unaligned proxy model. The objective is to iteratively adjust the proxy model so that it increasingly favors responses similar to those of the teacher while reducing its preference for its own initial outputs. To implement this, we employ the Direct Preference Optimization (DPO) algorithm \citep{rafailov2024direct}, which refines the proxy model by systematically preferring the teacher's responses.


Specifically, for a given input $x$, we iteratively sample a response $y$ from the teacher and $\hat{y}$ from the proxy. These responses form a preference pair $(x, y, \hat{y})$. To train the proxy model to prefer $y$ over $\hat{y}$, we define the following preference loss function:
\begin{equation}
{
\begin{aligned}
        & \mathcal{L}_{\text{DPO}}^{(i)}(x,y,\hat{y})= \\ 
        & \log{\sigma\left[\beta\log\frac{\pi_p^{(i)}(y|x)}{\pi_p^{(i-1)}(y|x)}-\beta\log\frac{\pi_p^{(i)}(\hat{y}|x)}{\pi_p^{(i-1)}(\hat{y}|x)}\right]},
\end{aligned}
}
\end{equation}
where $\pi_p^{(i-1)}$ is the proxy model from the previous training iteration. The overall preference loss over all the preference samples is defined as:
\begin{equation}
    \label{eq:online-preference}
        \mathcal{L}_{\text{Pref}}^{(i)} = \mathbbm{E}_{(x,y)\sim\mathcal{D}_p,{\hat{y}}\sim \pi_p^{(i)}(x)}\mathcal{L}_{\text{DPO}}^{(i)}(x,y,\hat{y}).
\end{equation}

At each iteration $i$, the proxy model is updated based on the combined objective that includes both the NLL loss and the preference loss:
\begin{equation}
    \label{eq:proxy-overall-objective}
    \mathcal{L}_{\text{Proxy}}^{(i)}=\mathcal{L}_{\text{Proxy-NLL}}^{(i)}+\mathcal{L}_{\text{Pref}}^{(i)}.
\end{equation}

This iterative process continues for a fixed number of iterations $k$ or until the proxy model converges.
Through this method, the proxy model $\pi_p$ is aligned to emulate the distribution of the black-box teacher $\pi_t$, becoming an effective intermediary for transferring knowledge to the student model.

\subsection{Knowledge Distillation}



To transfer knowledge from the black-box teacher to the student model $\pi_s$, we define the first training objective using teacher-generated sequences and the hard-label knowledge distillation objective:
\begin{equation}
    \mathcal{L}_{\text{Student-NLL}}=\mathbbm{E}_{(x,y)\sim \mathcal{D}_s}\left[-\log\pi_s(y|x)\right].
\end{equation}

Based on the proxy model aligned with the black-box teacher, which delivers accessible output distributions, we define another training objective for the student through soft-label knowledge distillation:
\begin{equation}
    \label{eq:student-Soft-KD}
    \begin{aligned}
        \mathcal{L}_{\text{Student-KL}}=\mathbbm{E}_{(x,y)\sim \mathcal{D}_s}\left[\mathbb{D}_{\text{KL}}(\pi_p(y|x)||\pi_s(y|x))\right].
    \end{aligned}
\end{equation}


In this process, the proxy model functions as an intermediary for the black-box teacher, facilitating the transfer of knowledge to the student model. However, as illustrated in Figure \ref{fig:match-mismatch} in Appendix, discrepancies between the teacher's and the proxy's output distributions persist even after aligning the proxy model, potentially degrading the effectiveness of knowledge distillation. To address these discrepancies, we propose a weighted approach to the soft-label knowledge distillation objective. By introducing weights, we dynamically adjust the influence of each sample based on the alignment quality between the proxy and the black-box teacher. This approach ensures that the student model prioritizes samples where the proxy's distribution closely matches the teacher's distribution and reduces focus on samples where it does not. The weights are calculated based on the log-likelihood of the teacher's output generated by the proxy, normalized by the mean and variance of these log-likelihoods:

\begin{equation}
    \begin{aligned}
        & w(x,y)=\sigma\left[\frac{\log\pi_p(y|x)-\mu}{\gamma}\right], \\
        & \mu=\mathbbm{E}_{(x,y)\sim\mathcal{D}_s}[\log\pi_p(y|x)], \\
        & \gamma^2 = \mathbb{V}\text{ar}_{(x,y)\sim\mathcal{D}_s}[\log\pi_p(y|x)],
    \end{aligned}
\end{equation}
where $w(x,y)$ is a weight reflecting the quality of the proxy’s prediction for the sample $(x,y)$, $\mathbb{V}\text{ar}(\cdot)$ is the variance operation, $\gamma$ is the standard deviation, $\sigma$ is the sigmoid function. Based on Equation (\ref{eq:student-Soft-KD}), we derive the sample-level weighted version of $\mathcal{L}_{\text{Student-KL}}$ as follow:
\begin{equation}
    \begin{aligned}
        & \mathcal{L}_{\text{Weight-KL}}= \\ 
        & \mathbbm{E}_{(x,y)\sim \mathcal{D}_s}\left[w(x,y)\mathbb{D}_{\text{KL}}(\pi_p(y|x)||\pi_s(y|x))\right].
    \end{aligned}
\end{equation}

Therefore, the overall objective for student knowledge distillation can be derived as:
\begin{equation}
    \label{eq:student-overall-objective}
    \begin{aligned}
    \mathcal{L}_{\text{Student}}=\mathcal{L}_{\text{Student-NLL}} + \alpha\mathcal{L}_{\text{Weight-KL}},
    \end{aligned}
\end{equation}
where $\alpha$ is a hyperparameter utilized to adjust the strength of the weighted KL loss.

This knowledge distillation strategy effectively blends the advantages of both black-box and white-box knowledge distillation methods, employing the proxy model to bridge the gap between black-box LLMs and open-source student LLMs.


\section{Experimental Setup}

In this section, we introduce the experimental settings of models, datasets, and method baselines.

\subsection{Models and Datasets}

\textbf{Teacher/Proxy/Student Models.} In Proxy-KD, we choose GPT-4 \citep{openai2023gpt-4} as the teacher, which is a powerful proprietary large language model. We select Llama-2-70b \citep{llama2} and Llama-2-13b \citep{llama3} as the proxy, respectively. Our student models come from two model types: Llama-1-7B \citep{llama} and Llama-2-7B \citep{llama2}.

\paragraph{Training Corpus. } We combine the OpenOrca \citep{openorca} and Nectar \citep{zhu2023starling} datasets as our training corpus, containing a total of 1M output sequences generated by the block-box teacher GPT-4. The OpenOrca dataset consists of instruction-following tasks, where GPT-4 is prompted to generate responses based on diverse input instructions. Nectar is a 7-wise comparison dataset, we filter and select those responses derived from GPT-4. 
Following \citet{li2024common}, we also incorporate synthetic data generated by GPT-4, based on existing benchmark training sets.
We split the original training corpus $\mathcal{D}$ into three parts: 10\% as $\mathcal{D}_w$ with 100K samples, 45\% as $\mathcal{D}_p$ with 450K samples, and 45\% as $\mathcal{D}_s$ with 450K samples.


\paragraph{Evaluation Benchmarks.} Evaluation benchmarks include complex reasoning dataset BBH \citep{suzgun2022challenging}, knowledge-based datasets AGIEval \citep{zhong2023agieval}, ARC-challenge \citep{clark2018think}, and MMLU \citep{hendrycks2021measuring}, commonsense reasoning dataset CSQA \citep{talmor-etal-2019-commonsenseqa}, and mathematical reasoning dataset GSM8K \citep{cobbe2021training}. 
All evaluated models apply a zero-shot greedy decoding strategy.

\begin{table*}[ht]
    \centering
    \vspace{-5mm}
    \begin{adjustbox}{width=0.98\textwidth}
        \begin{tabular}{l|c|c|cccccc|c}
        \toprule
        \textbf{Method} & \textbf{Student} & \textbf{Teacher / Dataset} & \textbf{AGIEval} & \textbf{ARC} & \textbf{BBH} & \textbf{CSQA} & \textbf{GSM8K} & \textbf{MMLU} & \textbf{Avg} \\
        \midrule
        \midrule
        \multicolumn{10}{c}{\small\textit{Black-Box Teacher}} \\
        \midrule
        GPT-4 & - & - & 56.40 & 93.26 & 88.0 & - & 92.0 & 86.4 & - \\
        \midrule
        \multicolumn{10}{c}{\small\textit{White-Box KD}} \\
        \midrule
        Forward KL \textsuperscript{$\clubsuit$} & Llama-1-7B & Llama-2-70B-Chat & 25.16 & 62.18 & 37.27 & 74.20 & 37.39 & 45.43 & 46.94 \\
        Forward KL \textsuperscript{$\clubsuit$} & Llama-2-7B & Llama-2-70B-Chat & 35.16 & 66.87 & 35.68 & 74.40 & 44.12 & {51.42} & 51.27 \\
        Forward KL \textsuperscript{$\clubsuit$} & Llama-2-7B & Llama-2-70B-Proxy & 35.56 & 69.34 & 45.72 & 74.97 & 46.34 & 51.13 & 53.84 \\
        MiniLLM \textsuperscript{$\clubsuit$} \citep{gu2023minillm} & Llama-2-7B & Llama-2-70B-Chat & 35.77 & 63.25 & 53.11 & 75.15 & 44.64 & 51.32 & 53.87 \\
        GKD \textsuperscript{$\clubsuit$} \citep{agarwal2023generalized} & Llama-2-7B & Llama-2-70B-Chat & 34.22 & 62.28 & 52.58 & 75.16 & 42.79 & 50.64 & 52.95 \\
        \midrule
        \multicolumn{10}{c}{\small\textit{Black-Box KD}} \\
        \midrule
        GPT-3 \citep{ho2022reasoning-teacher} & GPT-3-6.7B & text-davinci-002 & - & - & - & 56.76 & 6.75 & - & - \\
        FlanT5-XL \citep{fu2023specializing} & FlanT5-3B & text-davinci-003 & - & - & 39.0 & - & 22.4 & - & - \\
        FlanT5-XXL \citep{fu2023specializing} & Flant-11B & text-davinci-003 & - & - & 47.20 & - & 27.10 & - & - \\
        MCC-KD \citep{chen-etal-2023-mcc} & FlanT5-11B & ChatGPT & - & - & - & 84.93 & 33.99 & - & - \\
        MCC-KD \citep{chen-etal-2023-mcc} & Llama-1-7B & ChatGPT & - & - & - & 76.41 & 41.58 & - & - \\
        Orca-1 \citep{mukherjee2023orca} & Llama-1-13B & GPT-4 & 41.7	& 74.74 &	49.7 & 	- & 26.46 & 53.80 & - \\
        Orca-2 \citep{mitra2023orca} & Llama-2-7B & GPT-4 & 45.10 &	78.41 &	45.93 &	- & 47.23 & 53.70 & - \\
        Orca-2 \citep{mitra2023orca} & Llama-2-13B & GPT-4 & 49.93 & 83.36 & 50.18 & - & 59.14 & 57.73 & - \\
        WizardLM \citep{xu2023wizardlm} & Llama-2-13B & ChatGPT & 38.25&	74.74&	38.47&	-&	48.60&	55.00 & - \\
        Vicuna \citep{vicuna2023} & Llama-2-13B & ShareGPT & 29.3&	-&	23.3&	-&	-&	- & - \\
        Vanilla Black-Box KD \textsuperscript{$\clubsuit$} & Llama-1-7B & GPT-4 & 28.01 & 63.17 & 41.98 & 74.43 & 41.83 & 45.21 & 49.11 \\
        Vanilla Black-Box KD \textsuperscript{$\clubsuit$} & Llama-2-7B & GPT-4 & 34.71 & 66.85 & 46.68 & 74.43 & 49.51 & 49.82 & 53.66 \\
        \midrule
        \midrule
        TAKD \textsuperscript{$\clubsuit$} \citep{mirzadeh2019improved} & Llama-1-7B & GPT-4 & 25.73 & 63.61 & 38.87 & 73.01 & 39.45 & 39.12 & 46.63 \\
        TAKD \textsuperscript{$\clubsuit$} \citep{mirzadeh2019improved} & Llama-2-7B & GPT-4 & 35.05 & 67.18 & 43.0 & {76.04} & {47.54} & 48.09 & 52.82 \\
        Proxy-KD & Llama-1-7B & GPT-4 & 35.47 & 67.48 & 43.74 & 74.08 & 44.89 & 41.88 & 52.09 \\
        Proxy-KD & Llama-2-7B & GPT-4 & 36.59 & 71.09 & {53.40} & {75.18} & {53.07} & 51.35 & \textbf{56.78} \\
        \bottomrule
    \end{tabular}
    \end{adjustbox}
    \caption{Overall results on evaluated benchmarks. The superscript \textsuperscript{$\clubsuit$} represents our own implemented methods. Other results are from their original papers. All models utilize a zero-shot greedy decoding strategy for evaluation. Llama-2-70B-Proxy indicates that we use the aligned proxy as the white-box teacher for distillation.}
    \vspace{-0mm}
    \label{tab:main-results-1}
\end{table*}

\subsection{Training Configurations}

All experiments are conducted on 8×A100 Nvidia GPUs with 80GB memory. All proxy and student models are trained for only one epoch. We use a constant learning rate of 1e-5 and the Adam optimizer, with a max sequence length of 1024. 
We set hyperparamter $\alpha=100$ in Equation (\ref{eq:student-overall-objective}), and $k=16$ for the number of proxy alignment iterations.
All models are trained using LoRA \citep{lora} with mixed-precision: frozen parameters in bfloat16 and LoRA-trained parameters in float32. 

\subsection{Baselines}

We compare Proxy-KD with different white-box KD and black-box KD methods.

\textbf{White-Box KD.} For knowledge distillation with white-box teachers, we compare forward KL methods \citep{hinton2015distilling,agarwal2024policy} and reverse KL methods including MiniLLM \citep{gu2023minillm} and GKD \citep{agarwal2023generalized} (with the same hyperparameters set in the paper.). 
The chat version of Llama-2-70b is utilized as the white-box teacher. We also compare with using the aligned proxy as white-box teacher to perform distillation.

\textbf{Black-Box KD.} For knowledge distillation with black-box teachers, we compare the vanilla black-box KD methods \citep{mukherjee2023orca,mitra2023orca,xu2023wizardlm}, which directly fine-tunes the student on the data generated by the black-box teacher. We also compare Proxy-KD with the TAKD \citep{mirzadeh2019improved} method. 

For baselines implemented by us, we start from the same student checkpoint as Proxy-KD and use the same input prompts. In white-box KD, output sequences are generated by the white-box teacher, while in black-box KD, output sequences are generated by the black-box teacher.

\begin{table*}[t]
    \vspace{-5mm}
    \centering
    \renewcommand{\arraystretch}{1.2}
    \begin{adjustbox}{width=0.85\textwidth}
        \begin{tabular}{l|cccccc}
        \hline
         \textbf{Method} & \textbf{AGIEval} & \textbf{ARC} & \textbf{BBH} & \textbf{CSQA} & \textbf{GSM8K} & \textbf{MMLU} \\
        \hline
        \multicolumn{7}{c}{\small\textit{Studnet Model Distillation}} \\
        \hline
        $\mathcal{L}_{\text{Student}}$ & 36.59 & 71.09 & 53.40 & 75.18 & 53.07 & 51.35 \\
        \ \ w/o $\pi_p$ & 34.71  \textcolor{red}{(-1.88)} & 66.85   \textcolor{red}{(-4.24)} & 46.68  \textcolor{red}{(-6.72)} & 74.43  \textcolor{red}{(-0.75)} & 49.51  \textcolor{red}{(-3.56)} & 49.82  \textcolor{red}{(-1.53)} \\
        \ \ w/o $\mathcal{L}_{\text{Proxy}}$ & 35.05  \textcolor{red}{(-1.54)} & 67.18  \textcolor{red}{(-3.91)} & 43.0  \textcolor{red}{(-10.40)} & 76.04 \textcolor{blue}{(+0.86)} & 47.54  \textcolor{red}{(-5.53)} & 48.09  \textcolor{red}{(-3.26)}  \\
        \ \ w/o $\mathcal{L}_{\text{Pref}}$ & 35.38 \textcolor{red}{(-1.21)} & 66.11 \textcolor{red}{(-4.98)} & 52.51 \textcolor{red}{(-0.89)} & 75.51 \textcolor{blue}{(+0.33)} & 52.49 \textcolor{red}{(-0.58)} & 48.79 \textcolor{red}{(-2.56)}  \\
        \ \ w/o $\mathcal{L}_{\text{Weight-KL}}$ & 33.99  \textcolor{red}{(-2.60)} & 71.81 \textcolor{blue}{(+0.72)} & 51.50  \textcolor{red}{(-1.90)} & 75.11  \textcolor{red}{(-0.07)} & 52.91  \textcolor{red}{(-0.16)} & 49.47  \textcolor{red}{(-1.88)}  \\
        \hline
        \multicolumn{7}{c}{\small \textit{Proxy Model Alignment}} \\
        \hline
        $\mathcal{L}_{\text{Proxy}}$ & 49.12 & 87.67 & 66.04 & 82.18 & 78.24 & 68.62 \\
        \ \ w/o $\mathcal{L}_{\text{Pref}}$ & 48.31  \textcolor{red}{(-0.81)} & 86.93  \textcolor{red}{(-0.74)} & 62.16  \textcolor{red}{(-3.88)} & 80.95  \textcolor{red}{(-1.23)} & 79.15 \textcolor{blue}{(+0.91)} & 66.38  \textcolor{red}{(-2.24)}  \\
        \hline
    \end{tabular}
    \end{adjustbox}
    \caption{Ablation studies of Proxy-KD. We examine the impact of the proxy model $\pi_p$, proxy model alignment loss $\mathcal{L}_{\text{Proxy}}$, proxy preference loss $\mathcal{L}_{\text{Pref}}$, and weighted KL loss $\mathcal{L}_{\text{Weight-KL}}$ on the performance of the student model training, as well as the impact of the proxy preference loss $\mathcal{L}_{\text{Pref}}$ on the performance of the proxy model alignment.}
    \label{tab:ablation}
    \vspace{-0mm}
\end{table*}

\section{Result and Analysis}

In this section, we present the main results and additional experiments of Proxy-KD.

\subsection{Overall Results}

We show the comparison of Proxy-KD against baselines in Table \ref{tab:main-results-1}, the proxy models in Proxy-KD are based on Llama-2-70B backbone. Overall, the performance of black-box KD methods outperforms that of white-box KD methods, demonstrating the efficacy of distilling knowledge from powerful black-box models.

\textbf{Proxy-KD outperforms white-box KD and black-box KD methods.} Notably, Proxy-KD further enhances the performance, consistently achieving higher scores across most evaluated benchmarks compared to the white-box KD methods (e.g. MiniLLM and GKD) and the black-box KD methods. Improvement is particularly pronounced in the challenging datasets like BBH and GSM8K, where Proxy-KD obtains scores of 53.40 and 53.07, respectively, outperforming even larger models trained using traditional black-box KD methods. 

\textbf{Proxy-KD outperforms TAKD consistently.} TAKD performs even worse than vanilla Black-Box KD. When using Llama-1-7B as the student, vanilla Black-Box KD achieves an average of 49.11\%, while TAKD only reaches 46.63\%. Similarly, with Llama-2-7B as the student, vanilla Black-Box KD attains 53.66\% compared to TAKD’s average of 52.82\%. This decline in performance is likely due to TAKD's failure to account for the proxy alignment process, which is essential for effective closed-source KD. Introducing an unaligned proxy not only fails to enhance performance but actually degrades the performance of the student model.


\textbf{Proxy-KD outperforms white-box KD with an aligned proxy as the teacher.} Relying solely on an aligned proxy for white-box KD offers limited knowledge to the student. This suggests that the capabilities of closed-source teachers are more beneficial than those of open-source teachers, even after alignment, underscoring the superiority of distilling from closed-source LLMs.



We also present the performance changes of student models during the distillation process in Figure \ref{fig:main-results-1} in Appendix. We show the accuracy curves of students on the benchmark test sets for every 40K training steps. We compare three methods: vanilla black-box KD, Proxy-KD, and white-box KD (forward KL). 
The results show that Proxy-KD stands out with the most significant enhancements, indicating its superior capability to efficiently transfer the comprehensive knowledge of black-box teachers to student models. The steeper and more consistent improvement curves of Proxy-KD across benchmarks such as AGIEval, ARC, and particularly in complex tasks like BBH and GSM8K, underscore its robust and effective approach in leveraging proxy models for knowledge distillation. 

\subsection{Ablation Studies}

In this section, we examine the impact of different components within Proxy-KD. Llama-2-7B and Llama-2-70B are utilized as the backbones of the student and the proxy models, respectively.

\textbf{Effect of the Proxy Model.} 
The proxy model $\pi_p$ is crucial for the effectiveness of Proxy-KD. Removing the proxy model forces the distillation process to revert to hard-label knowledge distillation, leading to significant performance drops across multiple benchmarks: a decrease of 4.24 on ARC, 6.72 on BBH, and 3.56 on GSM8K, as shown in Table \ref{tab:ablation}. These declines underscore the proxy model's essential role in capturing and transferring the distributional knowledge from the black-box teacher, which is particularly important for tasks involving complex reasoning and mathematical challenges. Without the proxy, the student model fails to benefit from the detailed distributional guidance, resulting in markedly lower performance.

\textbf{Effect of Proxy Model Alignment.} 
The proxy model alignment, facilitated by the loss $\mathcal{L}_{\text{Proxy}}$, is vital for effective knowledge transfer. Table \ref{tab:ablation} shows that when the proxy is initialized directly from the Llama-2-70B checkpoint without alignment, the performance drops notably on BBH (-10.40), GSM8K (-5.53), and MMLU (-3.26). This decline illustrates the adverse effect of an unaligned proxy, which fails to approximate the teacher's distribution and consequently underperforms compared to models directly fine-tuned on teacher data. The slight increase on CSQA (+0.86) when skipping alignment might be attributed to the simplicity of the task, indicating potential overfitting to teacher outputs without proxy guidance. This reinforces the necessity of the alignment process to ensure the proxy effectively bridges the knowledge transfer from the black-box teacher to the student model across diverse and complex tasks.

\textbf{Effect of Preference Optimization.} 
Table \ref{tab:ablation} illustrates the significant role of preference optimization in enhancing the performance of both the proxy and student models. When the proxy preference loss $\mathcal{L}_{\text{Pref}}$ is removed, reducing the proxy alignment loss to $\mathcal{L}_{\text{Proxy-NLL}}$, we observe notable performance drops across various benchmarks. Specifically, the alignment of the proxy model with the black-box teacher deteriorates, as evidenced by decreases in scores on benchmarks like BBH and MMLU, which subsequently impacts the student model. The overall trend confirms that preference optimization is crucial for refining the proxy model's ability to emulate the teacher effectively.


\textbf{Effect of Weighted KL.} 
When $\mathcal{L}_{\text{Weight-KL}}$ is replaced with the standard KL loss $\mathcal{L}_{\text{Student-KL}}$, we also observe declines in performance across most benchmarks, indicating that the effectiveness of the distillation process diminishes. The results shown in Table \ref {tab:ablation} highlight that focusing on high log-likelihood distributions from the proxy, as facilitated by the weighted KL loss, significantly enhances the quality of knowledge transfer. The overall declines underscore that this weighting mechanism significantly improves the quality of knowledge distillation, enhancing the student's ability to learn from a well-aligned proxy.


\begin{figure}
    \centering
    \begin{adjustbox}{width=0.48\textwidth}
        \includegraphics{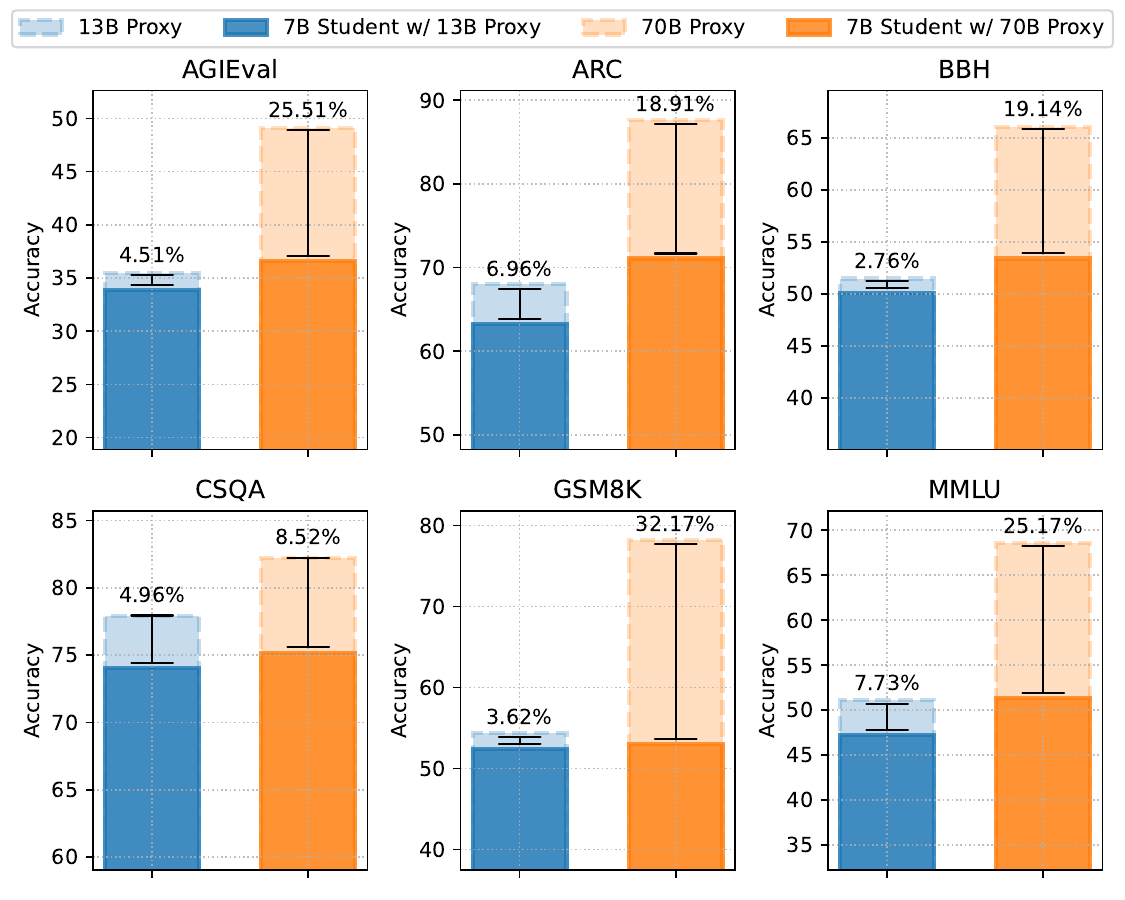}
    \end{adjustbox}
    \caption{Performance of student models under different proxy models. We also show the ratio of performance gap between the proxy models and the student models.}
    \label{fig:student-proxy}
    \vspace{-0mm}
\end{figure}

\subsection{Impact of Proxy Model's Capability}

How well the proxy aligned with the teacher can directly affect the performance of the student. The final alignment effectiveness of the proxy model depends on two factors: the design of the alignment algorithm and the inherent alignment capability of the proxy backbone model itself. In this section, we investigate the impact of the latter. We hypothesize that the size of the proxy model's parameters is crucial for its capacity to align with the black-box teacher's capability, especially when the teacher's parameter size is significantly larger than the proxy's. Experiments are conducted with Llama-2-70B and Llama-2-13B as the proxy backbone models. We show the performance of these aligned proxy models. As depicted in Figure \ref{fig:student-proxy}, the proxy model based on Llama-2-70B performs better than the one based on Llama-2-13B, the latter has fewer parameters. We also examine the impact of proxy models with different capacities on student performance. We observe that the stronger proxy based on Llama-2-70B yields better student performance than the weaker proxy based on Llama-2-13B. Furthermore, when using a proxy based on a backbone model with a larger capacity, the student demonstrates a greater potential for achieving higher performance.

\section{Conclusion}

This paper aims to tackle the challenge of knowledge distillation for black-box large language models (LLMs), where we can only access the outputs generated by the teacher model. Given the inaccessibility of the internal states of these black-box models, we introduce Proxy-KD, a novel approach that leverages a proxy model to enhance the distillation process. The proxy model is first aligned with the black-box teacher, closely mimicking its behavior. Then, the student model is trained using the combined knowledge from both the black-box teacher and the proxy model. Extensive experiments and analyses across a variety of well-established benchmarks demonstrate that Proxy-KD significantly outperforms existing black-box and white-box knowledge distillation methods.



\section*{Limitations}

The limitations of this work include the training time overhead associated with proxy model alignment, particularly when the proxy model has a large number of parameters. Additionally, the proposed preference optimization requires online sampling from the proxy model, further increasing the training time overhead. Another limitation is the type of experimental backbone models used. Due to resource constraints, this work only conducts experiments with the Llama model series, without including other model backbones such as Qwen \citep{bai2023qwen} or Mistral \citep{jiang2023mistral}.

\bibliography{custom}

\clearpage

\appendix

\section{Experimental Analysis}

\begin{table}[t]
    \centering
    \begin{adjustbox}{width=0.38\textwidth}
        \begin{tabular}{l|ccc}
        \toprule
        Models & \#GPUs & Hours/Round \\
        \midrule
        Llama-7B-SFT & 4 & 1.0 \\
        Llama-7B-Distill & 4 & 2.0 \\
        Llama-7B-GKD & 8 & 10.0 \\
        Llama-13B-SFT & 8 & 1.8 \\
        Llama-13B-Pref & 8 & 9.0 \\
        Llama-70B-SFT & 8 & 5.5 \\
        Llama-70B-Pref & 8 & 28.0 \\
        \bottomrule
        \end{tabular}
    \end{adjustbox}
    \caption{Training time overhead. We show the training hours per round for different methods. SFT is the supervised fine-tuning method, Distill is the knowledge distillation method, Pref is the preference optimization method. For GKD \citep{agarwal2023generalized}, student model is based on 7B, teacher model is based on 70B. Each round contains 40K training samples. }
    \label{tab:training-time}
\end{table}

\subsection{Analysis of Training Efficiency}

\begin{figure}
    \centering
    \begin{adjustbox}{width=0.48\textwidth}
        \includegraphics{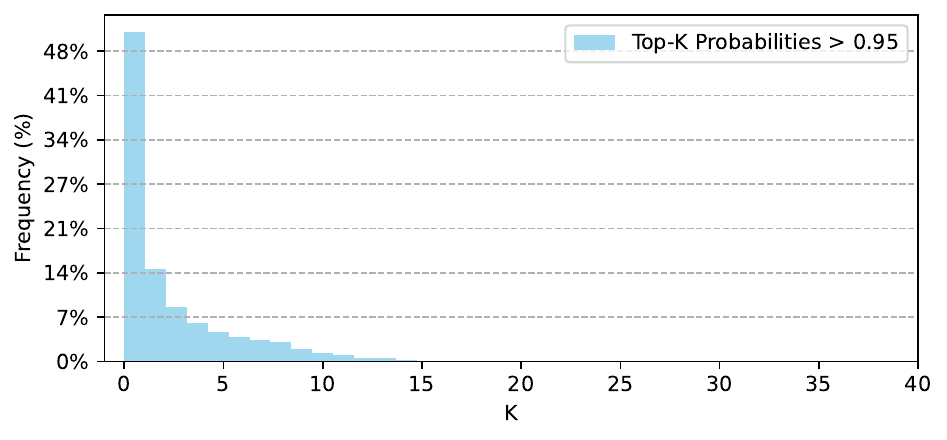}
    \end{adjustbox}
    \caption{The statistics of the cumulative probability within the Top K exceeding 0.95. The x-axis represents different values of K, while the y-axis shows the percentage of instances meeting this threshold.}
    \label{fig:top-k-tokens}
\end{figure}

We show the training time overhead for different methods in Table \ref{tab:training-time}. We show the training hours per round for supervised fine-tuning, knowledge distillation, and preference optimization methods across various model sizes. Each round contains 40K training samples. We note that preference optimization is the main time overhead due to online sampling from the proxy model.
In Proxy-KD, we obtain the proxy model's output distribution offline during student distillation. As Figure \ref{fig:top-k-tokens} shows, most probability mass is concentrated on a few tokens. To save memory, only the top 10 token indices and their logits are retained.

\subsection{Output Token Agreement}

\begin{figure}
    \centering
    \begin{adjustbox}{width=0.48\textwidth}
        \includegraphics{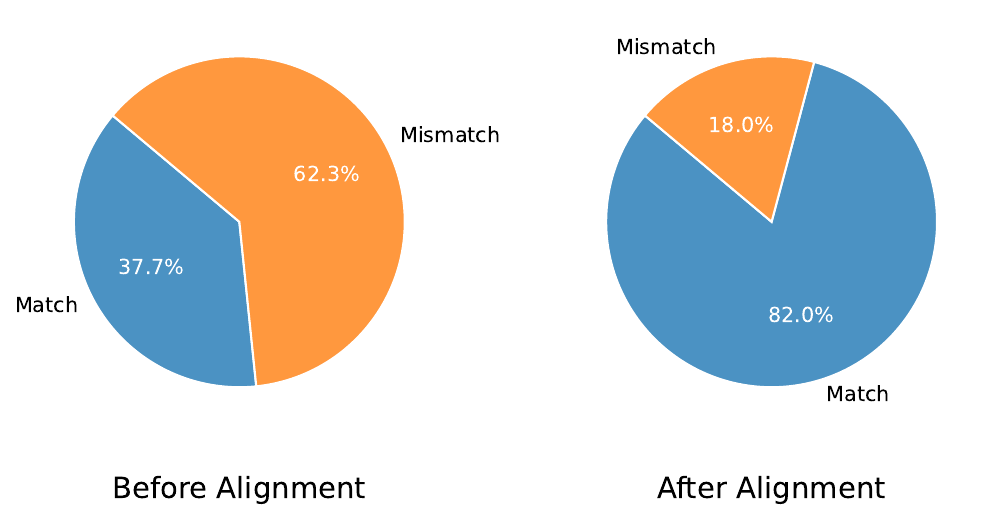}
    \end{adjustbox}
    \caption{The match ratio between the proxy and teacher's output tokens before and after alignment. If the top-1 token given by the proxy equals the token given by the teacher in a current step, it is considered a match; otherwise, it is considered a mismatch..}
    \label{fig:match-mismatch}
\end{figure}

To serve as a stand-in for the teacher model's output distribution, it's important for the proxy model's output to align with the teacher model's output distribution, which is achieved through proxy model alignment. We measure the change in agreement between the top-1 token given by the proxy and the token provided by teacher in current step, before and after alignment. 
To visualize this alignment, at each step, consider the top-1 token given by the proxy's output distribution and the token given by the teacher. If the top-1 token given by the proxy matches the token given by the teacher at the current step, it is considered a match; otherwise, it is considered a mismatch.
As shown in Figure \ref{fig:match-mismatch}, We find that after the proxy model alignment, the matched portions show a significant upward trend, indicating a trend towards alignment.

\subsection{Additional Results}

\begin{figure*}
    \centering
    \vspace{-0mm}
    \begin{adjustbox}{width=0.98\textwidth}
        \includegraphics{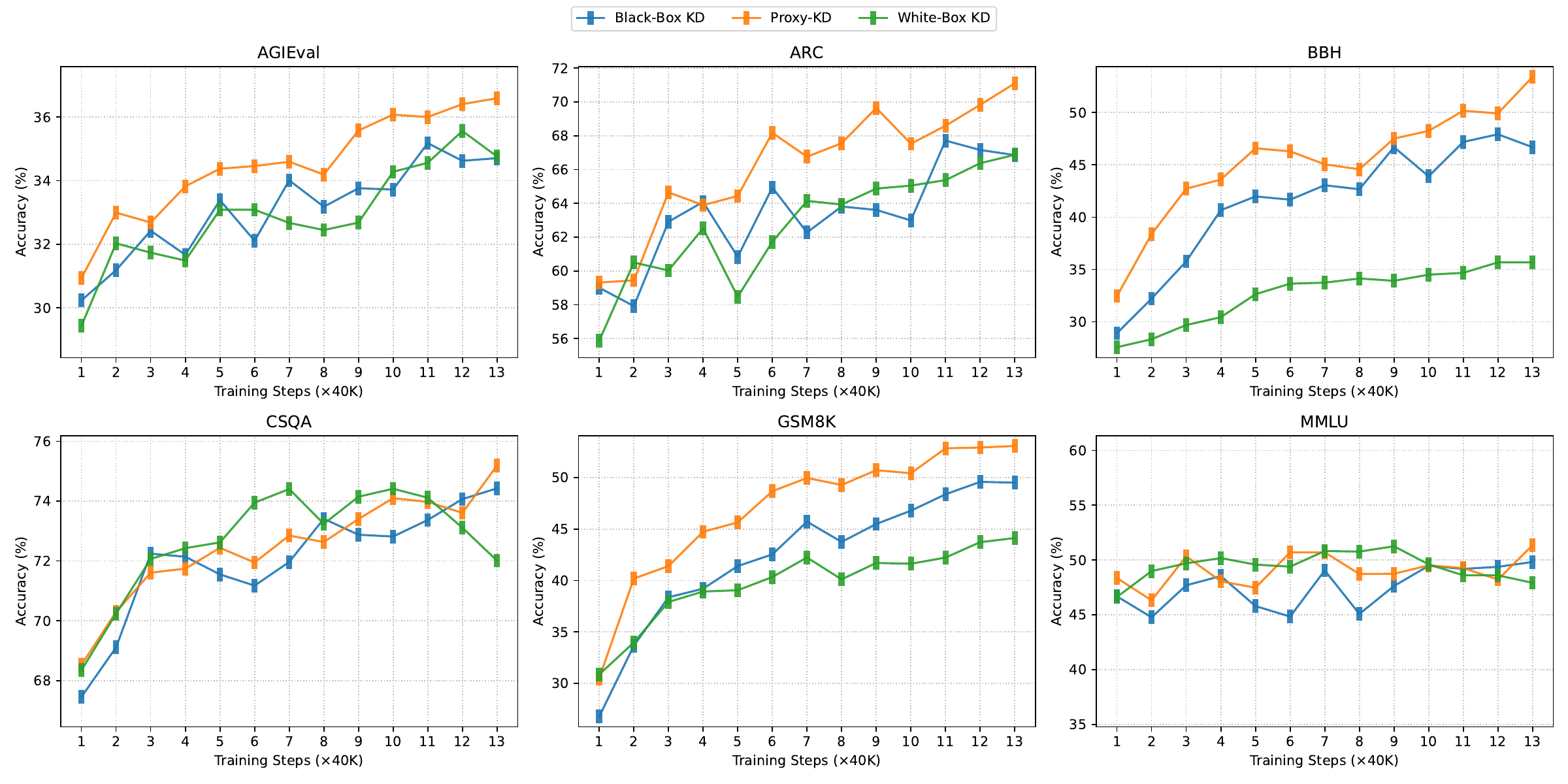}
    \end{adjustbox}
    \caption{Accuracy curves for student during distillation process. The y-axis is the accuracy on the benchmark test sets, and the x-axis is the number of training steps. We compare Proxy-KD with black-box KD (vanilla black-box KD) and white-box KD (forward KL) baselines. Notably, Proxy-KD did not show sign of saturation on some benchmarks, such as AGIEval, ARC, and BBH benchmarks.}
    \label{fig:main-results-1}
\end{figure*}

\begin{figure*}
    \centering
    \begin{adjustbox}{width=0.98\textwidth}
        \includegraphics{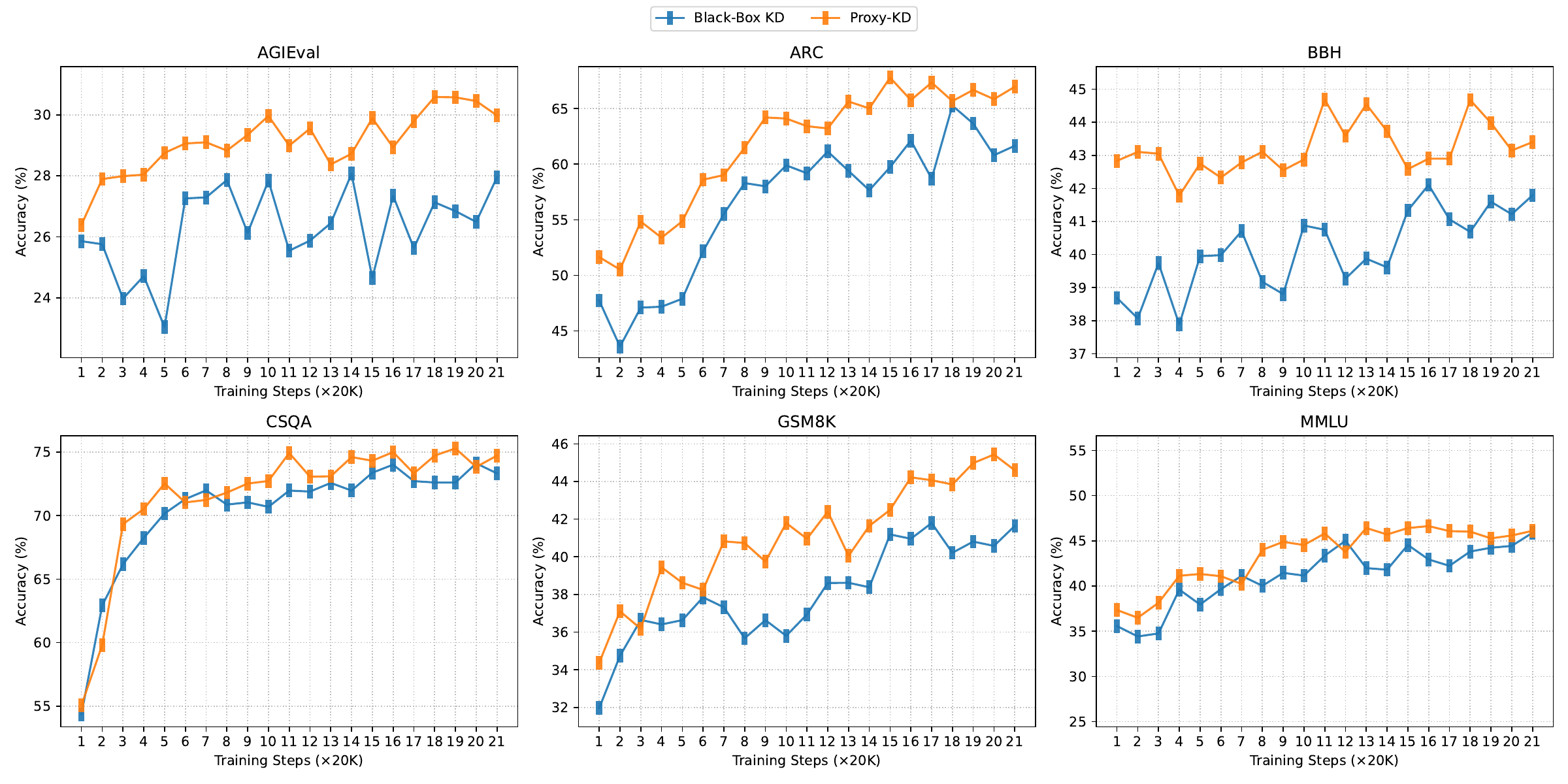}
    \end{adjustbox}
    \caption{Accuracy curves for student models during knowledge distillation process. The y-axis is the accuracy of students on the benchmark test sets, and x-axis is the number of training steps. We compare Proxy-KD with vanilla black-box KD. The students are based on Llama-1-7B, and the proxy is based on Llama-2-70B.}
    \label{fig:main-results-llama1}
\end{figure*}

We present the performance changes of student models during the distillation in Figure \ref{fig:main-results-1} and \ref{fig:main-results-llama1}. The student models are based on Llama-2-7B and Llama-1-7B backbone, and the proxy models are based on Llama-2-70B backbone. We test the accuracy of students on benchmarks for every 20K training steps. We compare Proxy-KD with vanilla black-box KD method and white-box KD method (Forward KL with Llama-2-70b-chat as white-box teacher) . We observe Proxy-KD consistently outperform vanilla black-box KD and white-box KD.

\end{document}